# Learn and Transfer Knowledge of Preferred Assistance Strategies in Semi-Autonomous Telemanipulation

Lingfeng Tao*, Michael Bowman*, Xu Zhou*, Jiucai Zhang^, and Xiaoli Zhang*

*Abstract* — Enabling robots to provide effective assistance yet still accommodating the operator's commands for telemanipulation of an object is very challenging because robot's assistive action is not always intuitive for human operators and human behaviors and preferences are sometimes ambiguous for the robot to interpret. Although various assistance approaches are being developed to improve the control quality from different optimization perspectives, the problem still remains in determining the appropriate approach that satisfies the fine motion constraints for the telemanipulation task and preference of the operator. To address these problems, we developed a novel preference-aware assistance knowledge learning approach. An assistance preference model learns what assistance is preferred by a human, and a stagewise model updating method ensures the learning stability while dealing with the ambiguity of human preference data. Such a preference-aware assistance knowledge enables a teleoperated robot hand to provide more active yet preferred assistance toward manipulation success. We also developed knowledge transfer methods to transfer the preference knowledge across different robot hand structures to avoid extensive robot-specific training. Experiments to telemanipulate a 3-finger hand and 2-finger hand, respectively, to use, move, and hand over a cup have been conducted. Results demonstrated that the methods enabled the robots to effectively learn the preference knowledge and allowed knowledge transfer between robots with less training effort.

## I. Introduction

TELEMANIPULATION is a branch of teleoperation [1] in which a human operator can remotely manipulate objects using the teleoperated robot's hands. Unlike other teleoperation branches such as teleapproaching and telefollowing, telemanipulation tasks require fine motion adjustments to grasp objects at specific angles or at specific points and to apply force in a particular manner. For example, remotely controlling a robot hand to plug a phone charger into a wall outlet involves strict motion constraints to prevent the robot finger from blocking the charger plug. Applications of telemanipulation include industrial inspection and repairing, space exploration, search and rescue, and assistive living robotics.

Most current telemanipulation approaches are master-slave control, where an operator's hands give motion commands through data gloves, optical tracking, or reflective markers, and a robot hand follows. Such approaches rely on the operator's cognitive spatial transformation reasoning and fine motion tuning to overcome the sense of disembodiment and the physical discrepancy [2][3] between the operator's hand and the robot's hand to satisfy the subtle motion constraints for task success. Indirect manipulation and visualization for complex telemanipulation tasks can impose a large physical and mental burden on the operator, increasing failure, and user frustration [4]. It usually takes hundreds of hours to adequately train an operator for a specific telemanipulation task and robot [5].

Current telemanipulation approaches still rely on the kinematic mapping between a human hand and a robot hand, whose performance is affected by the physical discrepancy (e.g., telemanipulate a 3-finger robot hand) and lack of consideration of task requirement/constraints. Existing efforts add passive constraints such as envelopes [6] and virtual fixtures [7] to the operation environment, which cannot achieve subtle motion adjustment in the manipulation process. Recent research has demonstrated that robots can blend human input with robot action by inferring human intent so it can provide more active assistance in teleoperation [8]-[10]. However, these methods are only implemented in teleapproaching using a robot arm with trajectory assistance, instead of controlling a robot hand for telemanipulation. Methods that enable robots to actively provide assistance to telemanipulate objects have not received enough research attention. To satisfy the fine motion requirement in telemanipulation, a robot also needs to

This material is based on work supported by the US NSF under grant 1652454. Any opinions, findings, and conclusions or recommendations expressed in this material are those of the authors and do not necessarily reflect those of the National Science Foundation.
*L. Tao, M. Bowman, X. Zhou and X. Zhang are with Colorado School of Mines, Intelligent Robotics and Systems Lab, 1500 Illinois St, Golden, CO 80401 USA (e-mail: tao@mines.edu, mibowman@mines.edu, xuzhou@mines.edu, xlzhang@mines.edu).
^J. Zhang is with the GAC R&D Center Silicon Valley, Sunnyvale, CA 94085 USA (e-mail: zhangjiucai@gmail.com).

understand the operator and provide active assistance yet still accommodate the operator's commands, that is, semi-autonomous telemanipulation is necessary.

Toward semi-autonomous telemanipulation, a critical problem to overcome is how to enable the robot to provide assistance that is preferred by the operator. Although active assistance is essential, it is unknown how much assistance is appropriate to balance task success with the operator's feeling of being in control. Due to the difference in hand structures, some motion assistance from the robot may surprise the operator with counterintuitive movements, which could introduce more burden to the human to correct the actions, reduce the operator's sense of system control and consequently increase the operator's resistance of using the robot. Although researchers can develop different control methods with the goal to improve the operation quality [11], the problem still remains in determining which control method is preferred by a human operator. To overcome these deficiencies, the robot needs to be equipped with preferred assistance knowledge to understand the operators' preferences on the manner of different assistance strategies. However, this understanding of human preferred assistance in telemanipulation has rarely been studied.

Learning preference needs human subject experiments, which normally present poor data quality caused by the ambiguity and uncertainty of human intent [12][13]. One issue for preference modeling is that the inherent data discrepancy caused by the human ambiguity and relatively small size of the human dataset may cause the training process to suffer from performance oscillation, convergence difficulty, overfitting problem, and converging to a local solution. Additionally, derived or learned models are mostly user-specific and robot-specific and cannot adapt to new users or robots. The problem remains of how robots can quickly learn useful assistance knowledge from the teleoperation scenario with human involvement and transfer the knowledge to other robots with less training efforts.

This paper provides a methodology for robots to learn the preferred assistance knowledge in a manner of the predicted rank of the available assistive robot control methods, which empowers a robot to choose the preferred method for flexible grasp generation that accommodates the operator's motion commands and at the same time autonomously regulate its pose to satisfy the operator's preferences (fig. 1). Such preferred assistance has the potential to reduce the frustration of the operator and help build better human-robot cooperation in telemanipulation tasks. The main contributions are as follows:

(1) *Preference-model-enabled assistance.* A preference model is developed to learn the preferred assistance knowledge, where the input is robot grasp configurations generated by control strategies, and the output is predicted ranking. An interpretation layer was designed to convert the raw input data to preference-related features based on domain criteria such as mimicking operator motion, maximizing task success, or minimizing travel distance. The preference model enables a robot to provide active assistance that is preferred by human operators.

(2) *Stagewise Preference Model Updating (SMU) methods.* To improve the stability while learning the preference model, we develop SMU methods with optimizing objectives: prediction accuracy (SMUPA), prediction error (SMUPE), and weights tendency (SMUWT) that update the model from model candidates stage by stage during the training process. The methods increase the stability of preference learning, reduce learning iteration, and improve the performance of the preference-aware models.

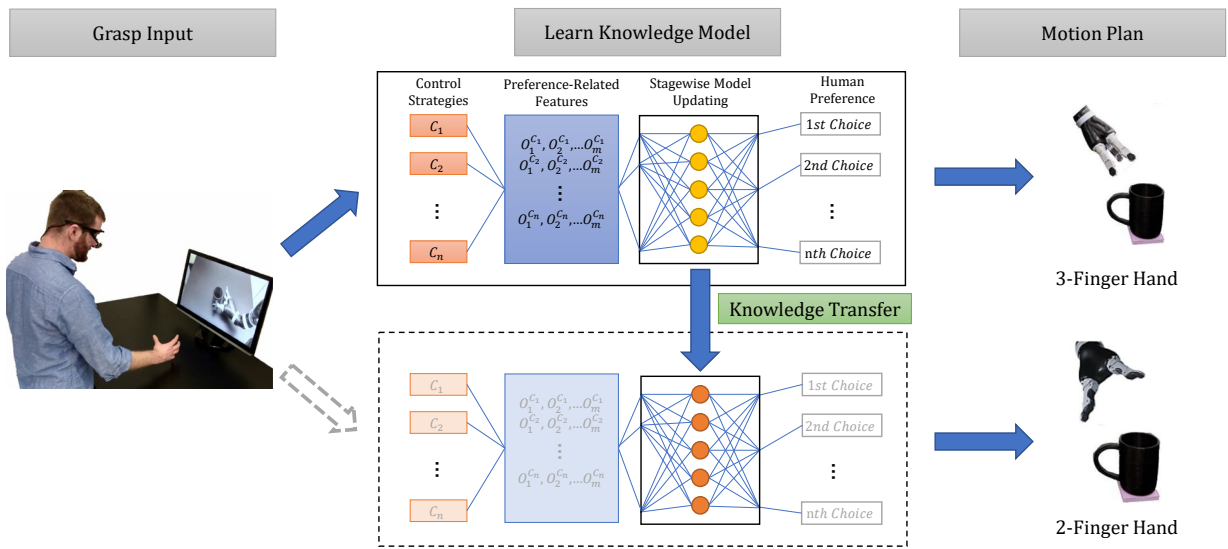

Fig. 1. The framework of modeling assistance knowledge and transferring knowledge between different robots. In a telemanipulation task, operator commands are processed by different control strategies to generate robot grasp poses, which will be converted to preference-related features that human may use to perceive the quality of robot assistance, such as mimicking human motion, following human intent and optimizing kinematics. A preference model is trained with the Stagewise Model Updating methods to overcome the inherent data imperfection caused by human ambiguity and to learn the relationship between human preference while controlling a 3-finger robot hand. Then we transfer the learned knowledge to a 2-finger hand with the modified knowledge transfer methods. The transferred model can be refined with much less training to achieve equivalent performance as if the preference model is specifically trained for the target robot.

(3) *Cross-robot knowledge transfer methods.* To avoid extensive robot-specific training, proactive knowledge transfer methods are developed to better extend the learned model across different robot hands. With the rank-based prediction of the preference model, one possibility is that the operator may rank the control methods differently for different robot structures. The goal of our knowledge transfer methods is to learn the accurate rank for the new robot structures with less training effort.

## II. RELATED WORKS

*1. Development in Telemanipulation*

Current telemanipulation research focuses more on analytical kinematic mapping methods based on the structures of the operator and robot [14]-[16]. Data-driven methods are widely used in kinematic mapping for 5-finger robots, such as end-to-end mapping for a humanoid robot hand, and these methods have good performance in mimicking human motion [17]. There are few applications for robotic end-effectors that differ from a human hand structure. Recent research indicates that bilateral telemanipulation that considers the specific kinematics of the devices involved which takes advantage of a virtual mediate object and forward and backward mapping algorithms can generate telemanipulation relation in asymmetric mapping [18]. But these methods still are pure kinematics and follow the perspective of the operator command; the robot lacks the ability to cooperate with and proactively aid the operator. For object manipulation tasks, task complexity and the additional requirement of fine motion operation require the robot to understand the operator's intent for task completion and preference for level of assistance, and autonomously regulate its configuration to ensure task success [19]-[22].

Reducing both the operator's workload and the difficulty of robot control through robot inference of operator intent for task completion is a recent topic in teleoperation, particularly in teleapproaching tasks. Research has demonstrated that in a target-approaching process, the robot agent can infer the target location by observing the operator's motion trajectory and provide motion assistance in approaching the target using linear blending strategies [23][24], virtual boundaries [25][26], and force guidance [27]. The bounded-memory adaptation model was used in human-robot mutual adaptation to predict the intent of an operator to maintain the operator's trust in teleapproaching a target object [28]. However, all these works only focused on teleapproaching an object not telemanipulating an object. For an object manipulation task, end-effector motion trajectory blending methods that treat the end-effector as a single point are not effective for robot finger control, because they rarely consider the fine motion constraints that are critical for the success of manipulation tasks. These subtleties in motion for object manipulation are also difficult to replicate with robotic hands because of their physical differences from human hands. We vision that different strategies will be developed to overcome these challenges in telemanipulation tasks. But, in practice, operators may have a preference for these strategies. The missing component is a model that can handle control strategy selection based on preferences. Therefore, there is a need to build preference models for telemanipulation.

*2. Development in Preference Modeling*

Modeling human preference is a trending topic in robotic research, especially in human-robot cooperation applications [29], such as, industrial assemblies, hybrid driving, and manipulation, in which a human and a robot need to cooperate with each other to complete a task. Human preference is a reference to understand how the action of the robot is perceived as effective or intuitive by the operator. A preference model helps to optimize the robot action and increase the trust and cooperation quality between humans and robots. Researches like [30] present a mathematical preference model based on probabilistic planning and game-theoretic algorithms to help the robot to understand and adapt to human preference in a leader-follower manner. The preference can also be learned from the online human trajectory demonstration for mobile manipulators such as assembly line robots [31]. Machine learning methods are getting popular recently. The approach in [32] formulates the model as a Markov Decision Process (MDP) and uses Inverse Reinforcement Learning (IRL) to learn human preference as a reward function. A similar approach in [33] also uses MDP modeling but learn the preference with regression-based and gradient-based methods. The sources of data are also expanded from human demonstration to subjective feedback like natural-language-facilitated preference learning [34]. The above approaches show that researchers have put great efforts into the preference modeling problem. However, these methods are applied in applications where both humans and robots can directly interact with the environment and have the ability to finish the task individually. The robot can autonomously execute action while taking human preference as a soft constraint. In telemanipulation, the unique problem is that the action of the robot is controlled by the human operator's command, which is a hard constraint that the robot must obey. The robot can only semi-autonomously assist with the consideration to balance the control performance and the operator's feeling of being in control. A preference model in telemanipulation is essential to help the robot to provide appropriate assistance. But, to the best of our knowledge, preference modeling in telemanipulation has been rarely reported.

### III. ASSISTANCE PREFERENCE MODEL

The human preferred assistance is defined to be those that are intuitive to human operators yet effective toward task success. The input of the preference model is characteristic raw data of robot hand configurations generated by different control strategies. These raw configuration data are converted to preference-related features by the model. The output of the model is based on human subjects' ranking of the preferred control strategies. The robot can use the learned model to determine the preferred control strategy (i.e., highest rank) to provide active assistance to achieve semi-autonomous telemanipulation. We define control strategies $C_n \in \mathcal{C}$, where $n$ represents strategy type. A control strategy type contains a group of controllers with different optimization criteria. For example, a control strategy like mimicking human motion can contain different controllers that emphasize the fingertip motion mimicking or joint motion mimicking or both. The optimization criteria of the controllers are used to design quantitative preference-related features $O_m \in \mathcal{O}$ that operators may use to perceive the quality of robot assistance. The raw grasp motion generated by each controller $C_n$ can be interpreted to preference-related features $C_n \rightarrow [O_1, O_2 \ldots, O_M]$. In Table I, an example list of potential preference-related features and corresponding controllers based on domain knowledge and literatures are shown.

### IV. STAGEWISE PREFERENCE MODEL UPDATING METHOD

A feed-forward neural network with sigmoid hidden neurons and linear output neurons is adopted for the preference model. Neural networks (NN) have been successful in supervised learning to map the relation between paired input-output training data [40]. A novel training method named stagewise model updating (SMU) can stabilize the training process and obtain the optimal model in a short training process. A stand-alone model $M_s$ is used during the training process and updated stage by stage. In each training episode, the snapshot model $M_s^j$ is saved in every $N$ iteration, which is then evaluated with the defined metric. The evaluation metrics include prediction accuracy (SMUPA)-based, prediction error (SMUPE)-based, and weights tendency (SMUWT)-based. At the beginning of the next training stage, the current model is updated with the best-performed snapshot model.

*1. SMU Evaluation Metrics*

**SMUPA:** Improving the prediction accuracy of the preference model is the priority during the training process. In this metric, the performance of the snapshot models is validated by checking their prediction accuracy based on the reranking of the raw prediction for each operator command, where the evaluation metric is computed as

$$A = \frac{1}{\Gamma}\sum_{\tau=1}^{\Gamma} C\big(s[M_s^j(\tau)] - r_\tau\big) \tag{1}$$

where $s[M_s^j(\tau)]$ is the reranked prediction of the controllers for one operator command, and $r_\tau$ is the true rank. Function $C$ compares the rank and output: if $s[M_s^j(\tau)] = r_\tau, C = 1$, if $s[M_s^j(\tau)] \neq r_\tau, C = 0$. $\Gamma$ is the number of testing data.

TABLE I
POTENTIAL PREFERENCE-RELATED FEATURES AND CORRESPONDING CONTROLLERS

| Controllers | Preference-related features | Description |
|---|---|---|
| Intent-based control [35] | $O_1 = \frac{1}{2}\sum_i^I (P_i(R) - T_i)^2$ | $T$ is the inferred human task intent. $P(R)$ are the probability distribution of each tasks given robot pose $R$. |
| Kinematics optimization [36] | $O_2 = \sum_i^I \frac{1}{\lambda_i}(R_i - H_i)^2$ | $R$ and $H$ are robot and human kinematics, include palm and finger configurations. $\lambda$ is the KL divergence for feature $i$. |
| Joint-wise optimization [37] | $O_3 = \sum_i^I (\vec{\rho}_i - \vec{\varphi}_i)^2$ | $\vec{\rho}$ and $\vec{\varphi}$ are robot and human joint configurations (Joint mapping is needed for discrepant structures). |
| Fingertip-wise optimization [38] | $O_4 = \sum_i^I (\vec{X}_i - \vec{Y}_i)^2$ | $\vec{X}$ and $\vec{Y}$ are robot and human fingertip locations (Finger mapping is needed for discrepant structures). |
| Vision-based optimization [39] | $O_5 = \sum_i^N \frac{1}{\psi_i}(I_{R,i} - I_{H,i})^2$ | $N$ is the number of pixels. $I_R$ and $I_H$ are processed pose images from robot and human. $\psi$ are the weights. |

**SMUPE:** This metric also focuses on improving the prediction accuracy of the model. The difference is that the performance is validated by comparing the cumulative RMS error between the actual rank and the raw prediction. where the error is computed as

$$E = \frac{1}{\Gamma}\sum_{\tau=1}^{\Gamma} \sqrt{\sum (M_s^j(\tau) - r_\tau)^2} \quad (2)$$

**SMUWT:** This metric assumes that the weights of the neuron should not change dramatically if the training process is stable and effective. The sudden change of the weights usually means the input data are insufficient and have large variances. To avoid the sudden change in weights and maintain a smooth performance increase, the metric is designed to monitor the tendency of the weights to change. When the change in weight is greater than a threshold, the SMUWT metric terminates the updates, recalls the last normal snapshot, and continues the training. The KL divergence [41] can determine how the weights change between the weights' distributions in the current model and snapshot model, which is calculated by

$$\mathcal{KL} = \sum_{k=1}^{K} P(w_k) \log\left[\frac{P(w_k)}{P(w'_k)}\right] \quad (3)$$

where $P(w_k)$ is the network weights distribution of the last updated model and $P(w'_k)$ is the network weights distribution of the current model, $k$ is the index of the neuron.

*2. Update Mechanism*

A threshold $\mathcal{T}$ is set to keep the training stable; it is tuned to reach the maximum training performance. When the KL divergence value exceeds $\mathcal{T}$, the current model will be replaced with the last saved safe model and will start a new training trial. When the KL divergence value does not exceed the threshold, the performance of the model may still decrease, and a validation step is used to avoid performance degradation. This validation is achieved by checking the primary goal of the prediction accuracy. If the prediction accuracy decreases, the current model will still be replaced with the last saved safe model. Because this method needs a reference model to calculate the KL divergence of the weight distribution, the algorithm kicks in at the second iteration.

In practice, the updates follow an ε-updating policy that updates the stand-alone model with randomly selected snapshots in probability of ε for exploration and updates the current model with the best-performed snapshot in the probability of $1 - \varepsilon$ for exploitation. The three strategies share a similar framework (fig. 2) but follow different updating laws.

V. TRANSFER PREFERRED ASSISTANCE KNOWLEDGE BETWEEN DIFFERENT ROBOTS

Conventional telemanipulation methods are robot-specific and task-specific, which require extensive efforts to generalize these methods. Our preference model that converts the raw data to preference-related features establishes a foundation for knowledge transfer across different robots. However, direct transfer of the reference model from one robot to another may not be a substantial approach, because the physical structure discrepancy of different robot hands causes value change to the feature spaces. For instance, a 2-finger hand may share a criterion such as task completion with a 3-finger hand because their structures are considerably different from a human hand; but a 5-finger hand and 4-finger hand may focus on mimicking human motion because their structures are more similar to a human hand.

*1. Positive Weights Transfer and Negative Weights Transfer*

The training process of the NN approximates the learning mechanism of biological neurons. In the learned NN of preferred assistance knowledge, the weights of each neuron record the knowledge that positively or negatively affects the human preference on the control strategies. When transferring knowledge between the robots, the fact is that the weights that store the knowledge are transferred. Inspired by the development of rectified linear unit (ReLU) layers [42], which is built on the neuroscience observation that control the firing rate of the total input current arising out of incoming signals at synapses [43], a knowledge transfer method is developed which modifies the weights with a positive rectifier function to transfer positive weights (TrPW) only, or with negative rectifier function to transfer negative weights (TrNW) only. The rectifier function allows the transferred knowledge to capture sparse representation, which is naturally suitable for human preference learning with sparse data.

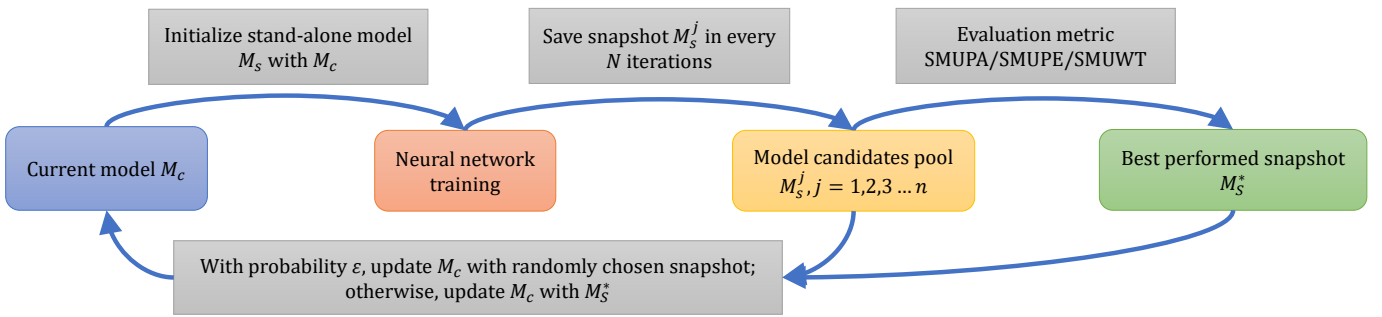

Fig. 2. The procedure of SMU strategies. First, the stand-alone model $M_s$ is initialized with the weights of the current model $M_c$. During the training, in every $N$ iterations, a snapshot model $M_s^j$ is saved to the model candidates pool. A designed evaluation metric is then used to evaluate the snapshot models in the pool; the current model $M_c$ will be updated with a randomly selected snapshot model in probability of $\varepsilon$ for exploration. Otherwise, the current model will be updated with the best-performed model $M_S^*$ for exploitation.

*2. Enhanced Weights Transfer*

The magnitude of the NN weights represents the contribution of an attribute of the input to the output. We hypothesize that the learned weights in the preference model are consistently distributed when transferring knowledge between similar robot hands. Thus, another knowledge transferring method, called enhanced weights transfer (TrEW), is developed to enhance the weights distributions. The weights are proportionally enhanced according to their distance to the average magnitude of all weights within corresponding hidden layers. The enhanced weight is calculated by

$$\theta' = (1 - \alpha)\theta - \alpha \frac{1}{L} \sum_{l=1}^{L} \theta_l \qquad (4)$$

Where $\theta$ is the weight, $\alpha$ is the gain of enhancement, $L$ is the index of weights in the layer.

## VI. EXPERIMENTS

*1. Experiment Setup*

Three control strategies were designed with selected preference-related features listed in Table I (the details are presented in the appendix). For simplicity, each strategy only contains one controller. The first is an intent-based strategy, which is designed to enable the robot system to understand the operator's task intent by reasoning the operator's motion and then generate its own motion to accomplish task success without explicit consideration of following the operator's motion. This strategy enables the robot to obey the *task* constraints without being interrupted by the physical discrepancy between the human hand and its own hand. For example, if the robot's task is to hold a cup for an individual to drink water, the robot's hand cannot cover the top of the cup. Also, for a robot to hand the cup to an individual, it is preferable to have the handle pointing out. The second strategy is a mimic-based strategy that makes the robot strictly follow the operator's motion commands using a fixed kinematic mapping policy. As motion commands may lie outside the bounds of the robot's capability, this strategy forces the robot to reach its physical limit but not attempting to use its own domain knowledge to explore a better alternative in these situations. The third strategy is an intent-mimic hybrid strategy, which determines the similarity of the operator command features to those known by the robot to find the level of importance they should have in the final grasp configurations. The importance is constructed as a penalty term into the formulation of the intent-based strategy. The new components added to the control system allow the robot to understand which attributes are common between itself and the operator as well as how similar these attributes are.

Human-involved experiments (fig. 3) were designed to collect training data for a robot to learn the preference knowledge and to validate our SMU strategies and knowledge transfer methods. We chose three principle tasks: Use, Handover, and Move. Each principle task consisted of 18 different human motion commands. For each human motion command, three different robot grasp motions were generated using the three designed control strategies for a 3-finger robot hand. To collect the human preference/value data, 20 human subjects were asked to rank the generated robot grasp for the 3-finger hand. The order of trials was randomly generated, the subjects were not told how any of the control strategies behaved, and the models were not explicitly marked with formulation names. In total, 1080 trials across 20 evaluators were collected. The SMU strategies were evaluated first to learn the preferred assistance knowledge for the 3-finger hand. The training was limited to 20 trials and each trial had 200 iterations, for a total of 4,000 training iterations. For each trial, the snapshot model during the training process was saved every 10 iterations; 20 models were saved in total. The baseline is a conventional supervised learning method, which trained the model with the same number of trials and iterations, but using a continuous training process that randomly chose 70% of the data for training, 15% of the data for validating, and 15% of the data for testing at each epoch. The learned models were transferred to the 2-finger hand to validate our knowledge transfer method, and then the SMU strategies were used to refine the transferred model.

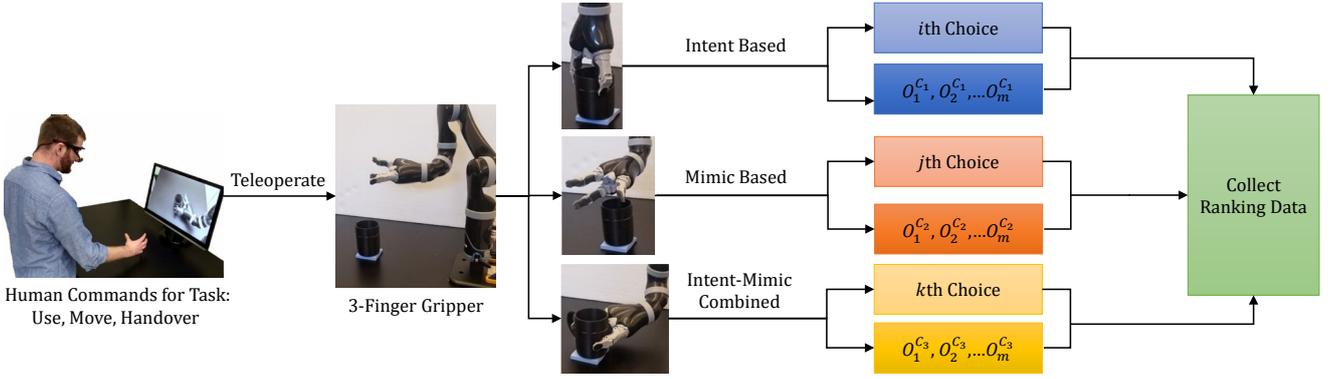

Fig. 3. Human-involved experiment for data collection. 54 commands are generated for principle task: usage, move, handover. A 3-finger hand is teleoperated with three control strategies: intent based, mimic based, intent-mimic combined. 20 evaluators gave rank for the three strategies. In total 1080 trials are collected.

*2. Experiment Evaluation Metrics*

The high variance and ambiguity in the collected preference data from the human subjects made it difficult to evaluate the performance of the trained preference model. For instance, different people may prefer different grasp configurations for the same manipulation task; consequently, evaluators may rank them in different orders, causing a high variance in the data. For example, one evaluator may have ranked the strategies [1, 2, 3], while another evaluator ranked them [2, 1, 3]. To deal with this issue in practice, we evaluated the performance of the learned model in a flexible way. Although a single control strategy does not satisfy all evaluators' preferences, certain control strategies are preferred than others. Like the previous example, strategies 1 and 2 are preferred, and strategy 3 is the least preferred. Thus, we formulate the evaluation criteria as a prediction problem to infer the control strategies with higher ranks when the operator telemanipulates the robot to complete a specific task. Instead of the winner-takes-all criterion, the model focuses on not only learning preferred control strategies but also understanding the least-preferred control strategies. This is useful knowledge for a robot as it attempts to understand human preferences and provide assistance that is aligned with those preferences. For example, a robot can provide the most preferred or sub-preferred choice and avoid the least-preferred choices.

## VII. RESULTS

*1. Assistance Preference Modeling*

Table II row 1 shows the prediction accuracy of the learned model. The average prediction accuracy for the 3-finger robot is 86.5%. From the model for each principle task, the highest prediction accuracy is 88.3% and the lowest prediction accuracy is 84.5%, which shows the consistency and feasibility of our methods for different tasks.

*2. Training with Stagewise Model Updating Methods*

Fig. 4 shows the training process of the preference model for all principle tasks while using the three SMU methods. Each data point represents the performance of the model after the training epoch. $\sigma^2_{SMU}$ and $\sigma^2_{Base}$ are the variance of prediction accuracy starting from the second data point. $D_{SMU}$ and $D_{Base}$ are the cumulative performance degradation in prediction accuracy when compared with that of the previous epoch. Overall, compared to the baseline methods across all tasks, the average performance of the SMUPA method is 77.6% more stable and reduce the performance degradation from 0.7448 to 0.2481; the SMUPE method is 8.5% more stable and reduces the performance degradation from 0.4185 to 0.2905; the SMUWT methods is 85.2% more stable and reduce the performance degradation from 0.6314 to 0. Specifically, the SMUPA method outperformed the baseline method in all tasks, but still experience performance oscillation in Hand Over and Move tasks. The SMUPE method performed well in Move and Use task but worse than the baseline in Hand Over task due to a massive performance drop at the late training stage. Among these three SMU methods, SMUWT can successfully maintain healthy updates in the neural network weights to avoid a performance drop caused by data disturbance. Even for the Move task where the other two SMU methods failed to keep the training stability and performance gain, the SMUWT method can still maintain the performance increase with stable training.

TABLE II
RESULTS OF MODEL LEARNING AND TRANSFERRING

|  | Hand Over | Move | Use | Average |
|---|---|---|---|---|
| 3-Finger (Learned Model) | 0.868 | 0.883 | 0.845 | 0.865 |
| 2-Finger (Directly Transfer) | 0.608 | 0.690 | 0.561 | 0.620 |
| 2-Finger (TrPW) | 0.642 | 0.771 | 0.689 | 0.701 |
| 2-Finger (TrNW) | 0.762 | 0.582 | 0.779 | 0.708 |
| 2-Finger (TrEW) | 0.712 | 0.711 | 0.599 | 0.674 |
| 2-Finger (Refined) | 0.894 | 0.872 | 0.847 | 0.871 |

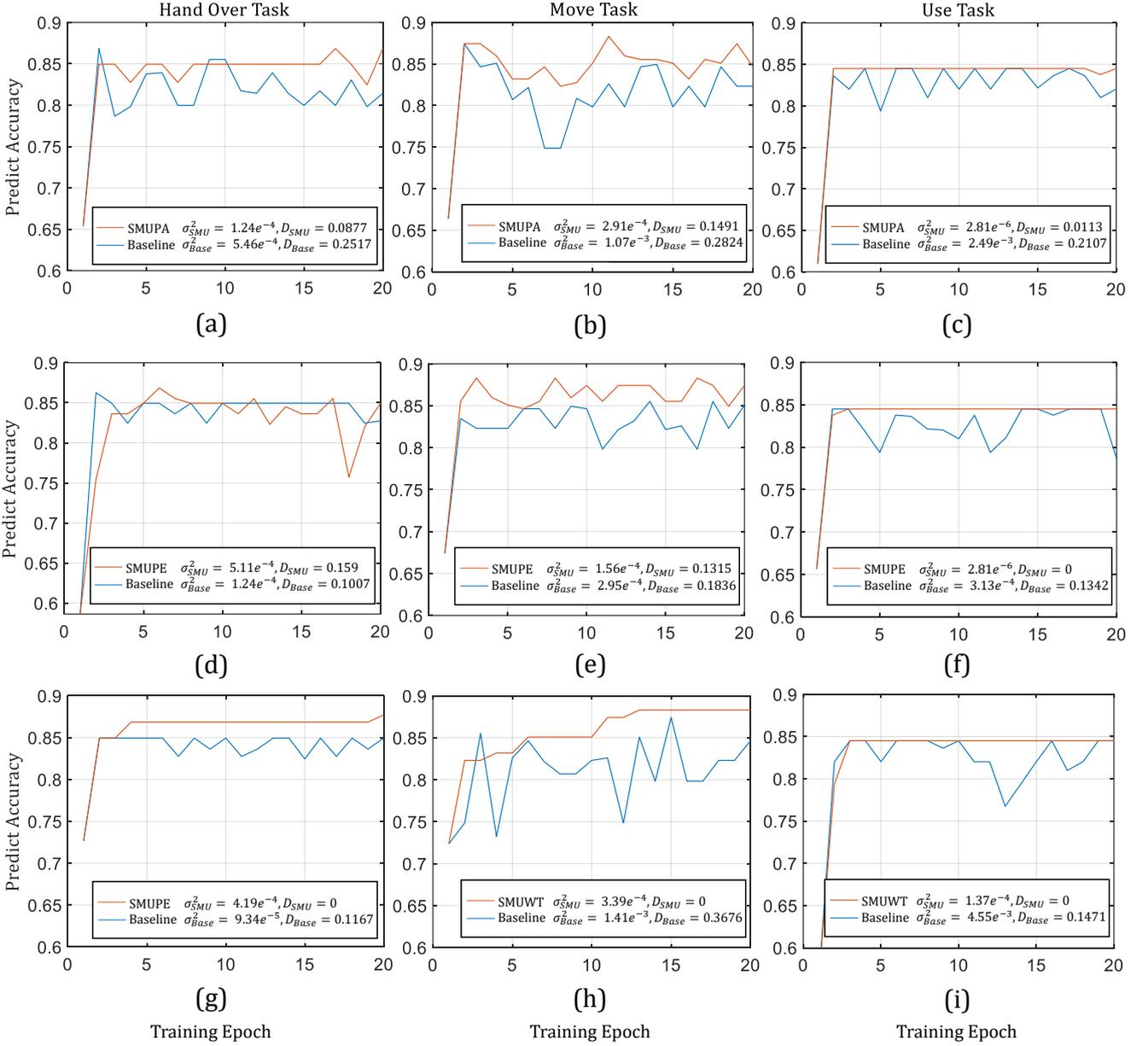

Fig. 4. The training process for each principle task (Hand Over, Move, Use) while learning the preference model with the three stagewise model updating strategies: SMUPA (a–c), SMUPE (d–f), and SMUWT (g–i). $\sigma^2_{SMU}$ and $\sigma^2_{Base}$ are the variance of prediction accuracy starting from the second data point. $D_{SMU}$ and $D_{Base}$ are the cumulative performance degradation in prediction accuracy during the training process.

The results of the learned preference model confirm that the operator's preference relates to the human motion command and the corresponding robot grasp configuration for a specific task. Comparing to the baseline, the average performance of all three SMU methods are 9.2% more stable and experience 47.4% less performance degradation in Hand Over task; 64.5% more stable and experience 67.5% less performance degradation in Move task, 86% more stable and experience 97.7% less performance degradation in Use task. Furthermore, the performance of the Move task is among the highest (accuracy = 0.883), the performance of the Use task is among the lowest (0.845), and the performance of the Hand Over task is in the middle (0.868). The Use task usually has more motion constraints than the other two tasks which may result in a clearer preference rank. However, the habitual differences of humans affect the data quality; for example, some people prefer holding the handle of a cup while drinking, while others prefer holding the body. These differences in preferences cause higher data variance and lower model performance, which caused the Use task with the baseline method to have the highest training instability and the lowest prediction accuracy. The results showed that the SMU methods were effective to handle this data variance to improve the stability of the learning process, and the Use task with the highest data variance was improved the most with 86% stability improvement.

*3. Transfer Preference Knowledge*

Row 2 of Table II presents the performance by directly transferring the learned model to the 2-finger hand. The average performance of all transferred preference models dropped to 0.62 compared to the original model. Rows 3 to 5 show the prediction accuracy of the transferred model while using different knowledge transfer algorithms. Statistically, the average prediction accuracy of all modified knowledge transfer methods is 0.694, which is 12% higher than direct knowledge transfer. Overall, the proposed methods outperform the direct transfer method in eight out of nine cases. The results for TrNW and TrPW methods show that the ReLU conversion is effective to transfer sparse information in most cases. The TrEW method had more performance degradation right after transfer than the other two methods. One reason is that when training with imperfect data, the contained noise, disturbance, and ambiguity are also learned in addition to the preferred assistance knowledge. Since we cannot identify which weight contains useful knowledge, this method may also enhance the learned imperfectness, which may reduce the performance. In summary, we can draw three main reasons for performance drops while transferring knowledge between different robots: (1) the physical discrepancy of the hand structure cause value change to the feature weights; (2) loss of information while transferring knowledge; and (3) the disturbance contained in the weights are also transferred. Thus, the transferred models need to be refined to resume the performance.

*4. Refine Transferred Model*

Row 6 of Table II shows the model performance after refining, the average prediction accuracy (0.871) shows the performance of the transferred model after refining can be comparable with if the model is specifically trained for the target robot. Fig.5 is an example of the refining process with three knowledge transfer methods for Move task and the SMUWT metric. The model transferred by the TrEW method reached the peak performance at the 2nd training epoch, which outperformed the other two transfer methods. A potential reason is that TrPW and TrNW models may need more data and training to refine because the hard zeros in the weights will affect the gradient backpropagation. Although TrEW has more performance degradation than the other two methods right after the transfer, it is much easier to refine to recover the performance because it transferred the complete distribution of the weighs which has less information loss and sets a good starting point while refining the transferred model.

Overall, the experimental results show that the combination of the TrEW method and the SMUWT strategy can provide the best performance concerning training stability, peak performance, and convergent speed. Fig.6 shows the example refining process for the SMUWT + TrEW method across three tasks. All cases reached peak performance in less than 4 training epochs with no degradation. In general, the refined models of the 2-finger robot hand for Hand Over task has the best prediction accuracy among the other two tasks. The preference model for Use task has slightly lower prediction accuracy than the other two tasks.

Fig.7 is an example of the performance changing flow chosen from the dataset when implementing the proposed methods for the Move task. The flow starts from the learned preference model for the 3-finger hand, which correctly matches the ground truth. When directly transferring the knowledge to the 2-finger hand, the model made a wrong prediction for all strategies. When using the TrEW knowledge transfer method, the performance increased compared to that of the direct transfer method. The model successfully predicted the second choice but confused the first and third choices. The model was then refined with the SMUWT methods in three training epochs with 540 samples (half of the data when train from scratch). The refined model successfully classified the first two preferred strategies and accurately identified the least preferred one. These results verify the feasibility of our assistance preference model and the assumption of knowledge transfer between different robots. It also proves the necessity of the proposed knowledge transfer methods and SMU methods.

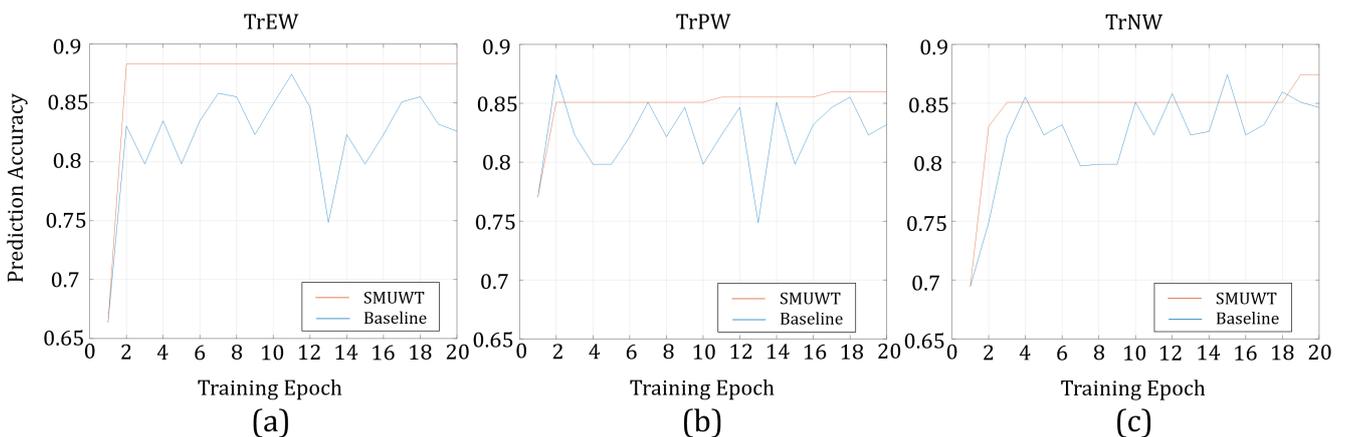

Fig. 5.   Refining process across the three transfer methods: (a) TrEW; (b) TrPW; (c) TrNW, for Move task, using SMUWT for best training performance.

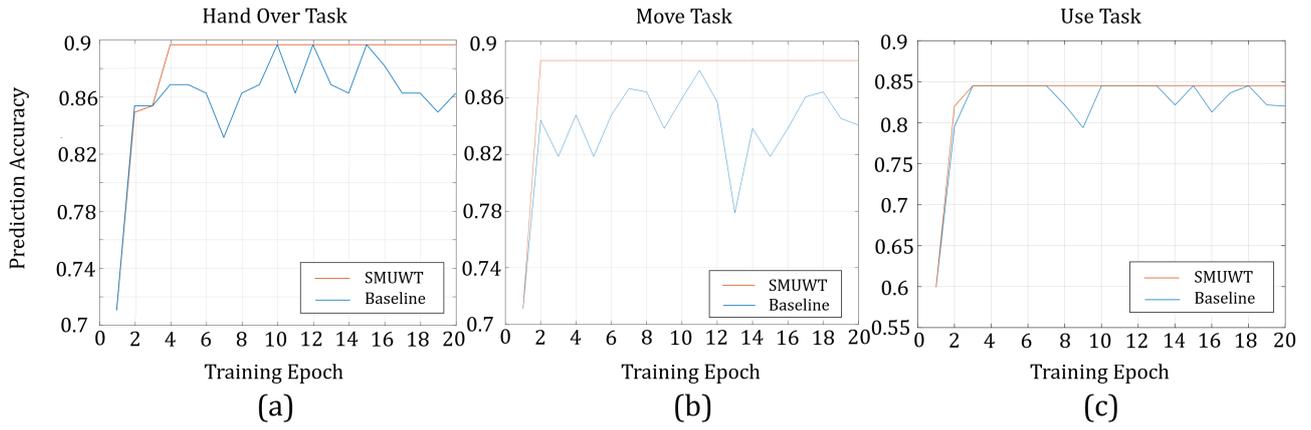

Fig. 6. Refining process across the three principle tasks: (a) Hand Over; (b) Move; (c) Use; with the model transferred with TrEW method, The SMUWT method was used to refine the model.

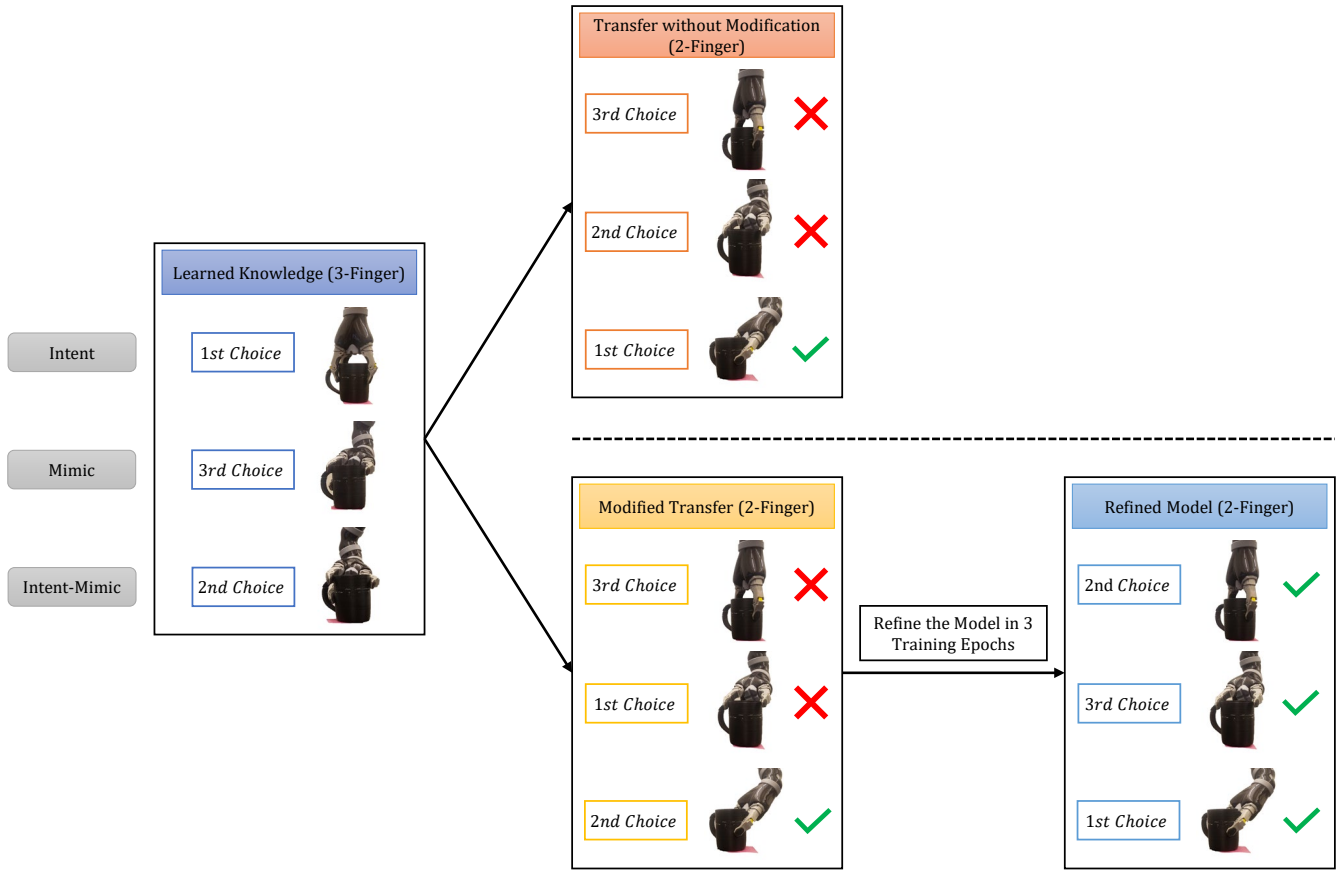

Fig. 7. An example of performance changing flow when applying the proposed methods. The learned preference model for 3-finger hand can make correct prediction. The performance dropped when exactly transfer the model to 2-finger hand. By implementing the modified knowledge transfer method and refining the transferred model in three training epochs, the model classifies the first two preferred strategies and accurately identifies the least preferred one for 2-finger hand.

## VIII. Discussion

*1. Use the Assistance Preference Model for Telemanipulation*

With the learned preference models, different preference-aware semi-autonomous manipulation schemes can be developed to enable robots to actively provide human-perceived effective assistance. For example, the robot can use the preference prediction to avoid the least preferred strategies and find the preferred assistance in the rest of candidates that maximize task reward, which is a more aggressive assistance scheme. The robot can also provide the first-ranked (most preferred) assistance to maximize operator preference, which can build better team cooperation but may not achieve the maximum task reward. Analysis of the confidence of the rank prediction and human adaptability can improve practicability when providing assistance in a real context.

*2. Applications for Other Robot Structures and Control Strategies*

Although for simplicity of testing and evaluation in this paper, we designed 3 control strategies and one controller for each strategy, the preference modeling and stabilized learning methods are expandable for a scenario with more control strategies and/or controllers and more preference-related features. The knowledge transfer methods can reduce the training efforts when adopting the reference models to different robot structures.

The difficulty of the knowledge transfer varies according to the level of difference between the source robot and the target robot. While transferring knowledge between different robots, the structures of the robots should be relatively similar; in our case, knowledge transfer between a 3-finger hand and 2-finger hand is applicable because the hands are similar, and all parameters are the same except for the number of fingers. Intuitively, it is more challenging to transfer the knowledge of a 2-finger hand to a 5-finger hand because their structures are more dissimilar, which may inhibit the sharing of transferable knowledge. For example, operators may prefer the mimic strategy more than other strategies when working with a 5-finger hand as it is more like the human hand. Thus, knowledge should be transferred between robots that are physically similar, like a 5-finger hand with 20 degrees of freedom to a 5-finger hand with 16 degrees of freedom, or a 5-finger hand to a 4-finger hand.

In general, after transfer and refinement, we expect the preferred rank to be similar but not necessarily identical. For example, for a 4-finger robot, the majority rank may be [intent-mimic, mimic, intent], and for a 5-finger robot the majority rank may be [mimic, intent-mimic, intent]. They still share transferrable knowledge to identify the first two preferred strategies, which are intent-mimic and mimic, and the least one intent-based control. Furthermore, if we have ten control strategies, and three of them are preferred by the operator, the rank of these three preferred strategies does not have to be the same. We expect that our model can identify the three preferred strategies and the learned knowledge can be transferred between different robots.

## IX. Conclusion

In this work, we developed a methodology for robots to choose the human-preferred way for providing a higher level of active assistance in semi-autonomous telemanipulation. We developed the preference models to learn the assistance preference knowledge in a manner of the predicted rank of the assistive robot control methods. We presented SMU methods to stably learn the preference model from ambiguous human preference data and different methods to transfer the preference model so different robots can use the model with fewer training samples. The experiment results demonstrated that the combination of the weights transfer method based on weight distribution (TrEW) and stagewise model updating strategy based on weights tendency (SMUWT) can implement the goal of knowledge transfer to reduce training efforts and ensure training stability. Our future research will concentrate on understanding the connection between the learned knowledge and the physical attributes of the task objects and subjects as well as developing and evaluating preference-aware assistance methods petitioned in discussion for telemanipulation with physical experiments.

XI. APPENDIX

The characteristic raw data is broken down into two categories: grasp attributes, and task attributes. Grasp attributes represent the hand kinematics, which include the palm orientation, palm center location, and finger configuration corresponding to the thumb and the index and middle fingers. We denote the set of robot grasp attributes as $\mathcal{R}$ and the set of human grasp attributes as $\mathcal{H}$. Task attributes $T$ describe tasks to be done.

*1. Intent-Based Strategy*

We denote the control variables of the robot as $R_a \in \mathcal{R}$. A set of $\{R_a\}$ produces a probability for each task, which is denoted as $P_b(R)$. An intent-uncertainty-aware human grasp model from previous work [44] is created to refer to the different task inference intents $T_b$. There are upper and lower bounds for model parameters, $U_a$ and $L_a$ respectively, which the robot must adhere to, such as physical limits of end effector position or joint angles, or force provided. We establish the intent inference, which consists of three principle tasks including Use, Move, and Hand Over. For example, for grasping a cup: Use is using or drinking from the cup, Move is moving the cup to another location, and Hand Over is handing the cup over to another agent. We use intent-uncertainty-aware human grasp model $\mathcal{M}$ to infer the intent $T_b$ in (5):

$$T_b = \mathcal{M}(\mathcal{H}) \quad (5)$$

The distribution $P_b(R)$ is used to quantify how much each task is satisfied by a given robot pose with features $R_a$. We use Naïve Bayes robot model $\mathcal{M}_r$ to produce the robot probability vector of satisfying the task $P_b(R)$ in (6) to (8), where $\mu_b$ is the average value for task $b$, $\Sigma_b$ is the covariance matrix for task $b$, and $d$ is the length of vector $R_a$.

$$P_b(R) = \mathcal{M}_r(R) \quad (6)$$

$$P(R_a|b) = \frac{1}{\sqrt{det(\Sigma_b)(2\pi)^d}} e^{-\frac{1}{2}(R_a-\mu_b)^T \Sigma_b^{-1}(R_a-\mu_b)} \quad (7)$$

$$P_b(R = R_a) = P(b|R_a) = \frac{P(R_a|b)P(b)}{\sum_b^B P(R_a|b)P(b)} \quad (8)$$

Upon developing the target probability vector and the robot probability vector, the intent-based strategy can be constructed based on the intent-based shared control criterion with added constraint, where the objective function is

$$C_1 = \min \left(\frac{1}{2} \sum_b (P_b(R) - T_b)^2\right)$$

$$s.t. \quad L_a \le R_a \le U_a \quad \forall_i$$

$$norm(R_a) = 1 \quad \forall_i \ needed \ for \ palm \ direction \quad (9)$$

*2. Mimic-Based Strategy*

If a human operator needs the robot to strictly follow the motion command, unintended errors may occur, but we can still achieve this goal by adding extra constraints to the intent-based strategy. The motion constraints can be explicitly dictated by adding the following set of constraints:

$$R_a = H_a \quad \forall_a \tag{10}$$

This will give the operator full control of all features of the robot. The new constraints added to the control diagram ensure the robot follows the human exactly by matching the robot features and human features to mimic the motion. The objective functions are

$$C_2 = \min\left(\frac{1}{2}\sum_b (P_b(R) - T_b)^2\right)$$
$$s.t. \quad L_a \le R_a \le U_a \quad \forall_a$$
$$norm(R_a) = 1 \quad \forall_a \text{ needed for palm direction}$$
$$R_a = H_a \quad \forall_a \tag{11}$$

*3. Intent-Mimic Hybrid Strategy*

We first define $\lambda_a$ as the KL divergence between the distribution of each feature.

$$\lambda_a = D_{KL}(\overline{\overline{R_a}}||\overline{\overline{H_a}}) = \ln\frac{\sigma_{H_a}}{\sigma_{R_a}} + \frac{\sigma_{R_a}^2 + (\mu_{R_a} - \mu_{H_a})^2}{2\sigma_{H_a}^2} - \frac{1}{2} \tag{12}$$

Additionally, the multivariate normal distribution between two populations can be used to determine the overall divergence between hand configurations in (13):

$$\gamma = D_{KL}(\overline{\overline{R}}||\overline{\overline{H}})$$
$$= \frac{1}{2}\left(\begin{array}{c} trace(\Sigma_H^{-1}\Sigma_R) + \\ (\mu_H - \mu_R)^T\Sigma_H^{-1}(\mu_H - \mu_R) - k + \ln\frac{|\Sigma_H|}{|\Sigma_R|} \end{array}\right) \tag{13}$$

The formulation results in making the mimic constraint from the previous formulation in the objective function to act as an elastic constraint which allows the robot to bend the rules on mimicking the human. The grasp position is generated by minimizing (14).

$$C_3 = \min\left(\frac{1}{2}\sum_b (P_b(R) - T_b)^2 + \frac{1}{\gamma}\sum_a^A \frac{1}{\lambda_a}(R_a - H_a)^2\right)$$
$$s.t. \quad L_a \le R_a \le U_a \quad \forall_a$$
$$norm(R_a) = 1 \quad \forall_a \tag{14}$$